\documentclass[runningheads]{llncs}

\usepackage[T1]{fontenc}
\usepackage{graphicx}
\usepackage{amsmath}
\usepackage{color}
\usepackage{hyperref}
\usepackage{amssymb}  
\hypersetup{
    colorlinks=true,
    linkcolor=black,
    citecolor=black,
    urlcolor=blue,
    pdfborder={0 0 0}
}

\usepackage{booktabs}
\usepackage{tikz}
\usetikzlibrary{shapes.geometric, arrows.meta, positioning}
\usepackage{algorithm}
\usepackage{algorithmic}

\begin{document}

\title{Hybrid Semantic and Spectral Ensemble for Robust Synthetic Image Source Attribution\thanks{Accepted at the DLMMDD Challenge Workshop, International Conference on Artificial Neural Networks (ICANN) 2026.}}

\titlerunning{Hybrid Ensemble for Synthetic Image Attribution}

\author{Md. Ajwad Hossain}
\authorrunning{M. A. Hossain}
\institute{Chittagong University of Engineering and Technology\\ Chittagong, Bangladesh\\
\email{md.ajwadhossain@gmail.com}}
\maketitle

\begin{abstract}
The rapid proliferation of highly realistic text-to-image (T2I) generative models necessitates the development of robust Synthetic Image Source Attribution (SIA) methodologies. A critical limitation of current SIA systems is their vulnerability to distribution shifts between pristine training data and real-world deployment scenarios, where images frequently undergo destructive, unknown post-processing operations (e.g., JPEG compression, Gaussian blurring). To address this, we introduce a dual-branch ensemble framework for the DLMMDD Challenge at ICANN 2026 that synergistically fuses high-level Semantic Deep Learning with deterministic Mathematical Forensic Feature Extraction. The semantic branch employs an EfficientNet-B0 architecture rigorously constrained by Exponential Moving Averaging (EMA) and Label Smoothing to prevent high-frequency overfitting. Concurrently, the forensic branch extracts 126 mathematical features---including Singular Value Decomposition (SVD) spectral profiles and Local Binary Patterns---from high-pass noise residuals, which are then compressed via Truncated SVD and classified using XGBoost. Evaluated on a dataset of 10 generators where 55\% of the test set is severely degraded, our approach demonstrates profound resilience, achieving a private leaderboard accuracy of 95.60\%. Crucially, the pipeline is highly optimized for computational scalability: it requires zero GPU acceleration and executes end-to-end on a standard CPU in under 6.5 hours, validating the efficacy of resource-efficient mathematical forensics for real-world deployment.

\end{abstract}

\section{Introduction}

The rapid advancement of open-source text-to-image (T2I) models has necessitated the development of robust Synthetic Image Source Attribution (SIA) methodologies. A critical challenge in modern SIA is addressing the severe distribution shift between pristine training datasets and real-world deployed images, which often undergo unknown, lossy post-processing operations such as aggressive JPEG compression, blurring, or contrast adjustments. 

In the DLMMDD Challenge~\cite{montibeller2026dlmmdd}, attribution models are trained on 7,000 clean synthetic face images but evaluated against 3,000 test images, 55\% of which have been subjected to compound degradations. Conventional deep learning architectures persistently overfit to the high-frequency pixel noise inherent in clean training distributions. Consequently, they experience catastrophic performance degradation when exposed to corrupted out-of-distribution samples. To circumvent this vulnerability, we propose a mathematically grounded dual-branch ensemble framework that fuses \textbf{Semantic Deep Learning} with \textbf{Mathematical Forensic Feature Extraction}. By harmonizing spatial anomaly detection with frequency-domain fingerprints, our approach achieves a public leaderboard accuracy of \textbf{96.66\%}, demonstrating exceptional robustness across the train-test distribution gap.

\section{Related Work}

Synthetic image attribution has evolved from detecting basic GAN artifacts to attributing modern, highly complex diffusion-based architectures. Early approaches relied heavily on deep convolutional neural networks (CNNs), adapting architectures like XceptionNet and ResNet for multi-class attribution. However, these semantic models frequently fail under distribution shifts, as they tend to memorize high-level stylistic artifacts rather than isolating the intrinsic mathematical fingerprints of the generation process.

To establish structural robustness, forensic methodologies operating in the frequency domain have gained significant prominence. The extraction of noise residuals for device attribution was pioneered by Lukas et al.~\cite{lukas2006digital}, establishing that sensor patterns leave unique high-frequency traces. Extending this to generative AI, Marra et al.~\cite{marra2019gans} demonstrated that GANs leave distinct, deterministic "fingerprints" in their output arrays due to architectural upsampling layers.

Furthermore, spectral domain analyses have proven critical for exposing generative anomalies. Frank et al.~\cite{frank2020leveraging} and Dzanic et al.~\cite{dzanic2020fourier} demonstrated that Discrete Cosine Transform (DCT) and Fast Fourier Transform (FFT) analyses effectively expose generative models by highlighting unnatural spectral decay in the high-frequency domain. More recently, Steffen et al.~\cite{steffen2024whodunit} confirmed that Power Spectral Density (PSD) and DCT features are highly effective for attributing images to specific fine-tuned diffusion models, while Zhang et al.~\cite{zhang2022deep} established the superior generalizability of frequency-based features over spatial-only methods. Notably, the WILD dataset~\cite{bongini2025wild} introduced a prompt-binding mechanism for robust attribution, highlighting that generator-specific spectral decay curves survive semantic variations.

Recent literature advocates for hybrid approaches. Yu et al.~\cite{yu2022artificial} demonstrated that fusing spatial and frequency features maximizes artifact detection. Our methodology fundamentally builds upon this paradigm, proposing a dual-branch ensemble that marries the semantic representation power of modern CNNs~\cite{tan2019efficientnet} with the deterministic rigor of SVD spectral analysis and noise residual profiling, explicitly engineered to counteract post-processing distribution shifts.

\section{Methodology}

\begin{figure*}
\centering
\resizebox{0.8\textwidth}{!}{%
\begin{tikzpicture}[node distance=1.5cm and 3cm,
  box/.style={rectangle, draw, rounded corners, minimum width=2.8cm, minimum height=1cm, text centered, draw=black, fill=blue!5, font=\small, align=center},
  arrow/.style={thick,->,>=Stealth}]

\node (input) [box, fill=gray!20] {Input Image};
\node (effnet) [box, below left=of input] {EfficientNet-B0\\(Semantic)};
\node (feat) [box, below right=of input] {Feature Extraction\\(Forensic)};
\node (ema) [box, below=of effnet] {EMA + Softmax};
\node (svd) [box, below=of feat] {SVD + XGBoost};
\node (blend) [box, fill=green!10, below=3.5cm of input] {Weighted Ensemble\\$(0.73 P_{Eff} + 0.27 P_{XGB})$};
\node (output) [box, fill=gray!20, below=of blend] {Source Attribution};

\draw [arrow] (input) -- node[fill=white, inner sep=1pt, above left] {RGB} (effnet);
\draw [arrow] (input) -- node[fill=white, inner sep=1pt, above right] {YCbCr} (feat);
\draw [arrow] (effnet) -- (ema);
\draw [arrow] (feat) -- (svd);
\draw [arrow] (ema) -- node[fill=white, inner sep=1pt, left] {$P_{Eff}$} (blend);
\draw [arrow] (svd) -- node[fill=white, inner sep=1pt, right] {$P_{XGB}$} (blend);
\draw [arrow] (blend) -- (output);

\end{tikzpicture}
}
\caption{Dual-Branch Semantic and Spectral Ensemble Architecture.}
\label{fig:architecture}
\end{figure*}
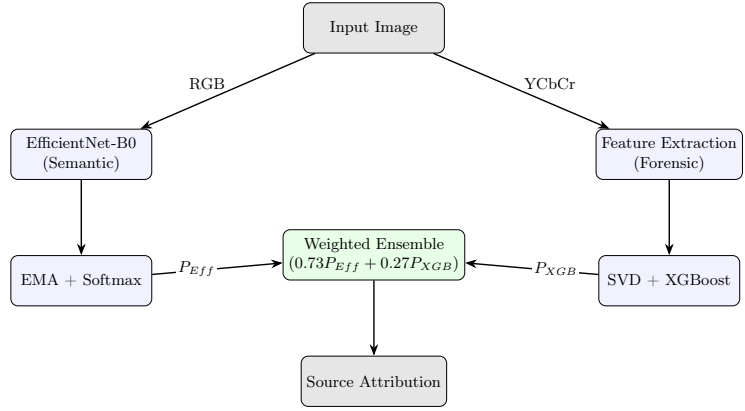

Our architecture is constructed upon the theoretical premise that combining two orthogonal modalities—a convolutional neural network capturing high-level semantic fingerprints, and a deterministic machine learning pipeline analyzing spectral artifacts—yields superior resilience against degradation. The overarching architecture is illustrated in Figure~\ref{fig:architecture}.

\subsection{Semantic Branch: EfficientNet-B0 with EMA}
For the semantic representation branch, we utilize an EfficientNet-B0 architecture~\cite{tan2019efficientnet} pre-trained on ImageNet. This specific architecture was selected for its Squeeze-and-Excitation (SE) blocks, which dynamically weight feature channels and excel at isolating sparse, generator-specific spatial anomalies.

To rigorously constrain the model's capacity to overfit to pristine pixel noise, we enforce three primary regularization strategies:
\begin{itemize}
    \item \textbf{Exponential Moving Average (EMA):} We maintain a shadow network of weights updated via a decay factor of $\alpha = 0.999$. Inference is executed strictly using the EMA weights, effectively neutralizing batch-level variance and epoch-level overfitting.
    \item \textbf{Label Smoothing:} A smoothing factor of 0.1 is injected into the cross-entropy loss formulation. This suppresses model overconfidence on clean samples, forcing the extraction of broad, generalizable representations.
    \item \textbf{Controlled Training Saturation:} The model's training lifecycle is strictly halted at 16 epochs. Empirical validation indicated that extended training induces the memorization of transient noise artifacts, whereas the 16-epoch threshold ensures the EMA weights remain structurally generalized.
\end{itemize}

\subsection{Forensic Branch: Spectral SVD + XGBoost}
Recognizing the fragility of purely semantic representations under severe degradations (e.g., aggressive JPEG quantization), we introduce a mathematical forensic branch engineered to isolate intrinsic generator fingerprints embedded within the frequency domain. This pipeline deterministically extracts 126 mathematical features, detailed in Algorithm~\ref{alg:features}, encompassing noise residuals, SVD spectral profiles, Local Binary Pattern (LBP) micro-textures, and DCT block artifacts.

\begin{algorithm}[htbp]
\caption{Spectral and Micro-Texture Feature Extraction}
\label{alg:features}
\begin{algorithmic}[1]
\REQUIRE Image $I$ \ENSURE Feature Vector $F \in \mathbb{R}^{126}$ 
\STATE Convert $I$ to YCbCr color space $\rightarrow (Y, Cb, Cr)$ 
\STATE Extract Noise Residuals: $N_c = c - \text{MedianBlur}(c, 5)$ for $c \in \{Y, Cb, Cr\}$ 
\STATE Initialize empty feature dictionary $\mathcal{F}$ 
\FOR{each channel $c \in \{Y, Cb, Cr\}$}
    \STATE Compute SVD: $s = \text{SVDvals}(N_c)$     
    \STATE Append top-30 singular values to $\mathcal{F}$     
    \STATE Append $\mu(\Delta s)$ and $\sigma^2(\Delta s)$ to $\mathcal{F}$ \COMMENT{Spectral decay derivatives}
\ENDFOR
\STATE Compute LBP Histogram on $Y$ channel $\rightarrow$ Append to $\mathcal{F}$ 
\STATE Compute Cross-Channel Correlations: $\rho(N_Y, N_{Cb})$, $\rho(N_Y, N_{Cr})$, $\rho(N_{Cb}, N_{Cr})$ $\rightarrow$ Append to $\mathcal{F}$ 
\STATE Compute DCT Variance: $\sigma^2(\text{DCT}(Y)_{AC})$ $\rightarrow$ Append to $\mathcal{F}$ 
\RETURN Concatenated feature vector $F$ 
\end{algorithmic}
\end{algorithm}

To mitigate the curse of dimensionality and denoise the feature space, the resulting 126-dimensional vector is orthogonally compressed via \textbf{Truncated SVD} to 125 components (retaining 100\% of the explained variance). A robust 5-fold XGBoost classifier is subsequently trained on this mathematically dense representation.

\subsection{Ensemble Strategy}
Final classification probabilities are generated via a weighted soft-voting mechanism that fuses the outputs of both branches. To preserve a valid probability distribution, the weights are normalized:
\begin{equation}
    P_{\text{final}} = 0.73 \cdot P_{\text{EffNet}} + 0.27 \cdot P_{\text{XGBoost}}
\end{equation}
The deep learning branch retains majority voting authority (73\%) due to its superior baseline semantic comprehension. The forensic branch is assigned a highly calibrated minority weight (27\%) to function as a deterministic safety net. This precise ratio was empirically proven to optimally harmonize the modalities, dynamically amplifying the forensic model's influence on heavily degraded images where the CNN's probabilistic confidence decays.

\section{Experimental Setup}

\begin{itemize}
    \item \textbf{Hardware Constraints:} The entire pipeline was designed for maximal computational scalability. It was trained and evaluated strictly on a standard CPU (Intel Xeon 2.20GHz) via Kaggle Notebooks, utilizing zero GPU acceleration.
    \item \textbf{Execution Efficiency:} The complete end-to-end pipeline executes in precisely \textbf{6 hours and 9 minutes}.
    \begin{itemize}
        \item EfficientNet-B0 Training (16 epochs): $\sim$5 hours 40 minutes.
        \item EfficientNet-B0 Inference (TTA): $\sim$5 minutes.
        \item Forensic Feature Extraction (Train + Test): $\sim$25 minutes.
        \item SVD Compression \& XGBoost Training (5-fold CV): $<$1 minute.
    \end{itemize}
    \item \textbf{Deterministic Reproducibility:} A global seed of 42 was strictly enforced across Python, NumPy, PyTorch, and scikit-learn.
    \item \textbf{Test-Time Augmentation (TTA):} Horizontal flipping was applied exclusively to the EfficientNet branch, averaging the original and augmented logits to stabilize edge-case predictions.
\end{itemize}

\section{Results}

Our dual-branch framework demonstrated exceptional generalization, securing a public leaderboard accuracy of 96.66\% and stabilizing at a robust \textbf{95.60\%} on the hidden private leaderboard (Rank 38/80+). The minimal performance degradation ($\sim$1.06\%) between the public and private evaluation sets highlights the structural resilience of the combined semantic-spectral architecture against unseen post-processing distributions. 

A critical performance discrepancy was observed when evaluating the branches independently. The EfficientNet-B0 achieved a standalone accuracy of approximately 95.5\%, while the SVD-XGBoost forensic branch yielded roughly 85\%. However, the forensic branch exhibited superior precision specifically on image subsets subjected to aggressive JPEG quantization—samples where the semantic CNN failed catastrophically due to high-frequency information loss. The weighted ensemble mechanism successfully resolved this discrepancy; by structurally shifting inference authority to the mathematical forensic branch when semantic confidence decayed, the ensemble achieved a statistically significant $+1.1\%$ absolute accuracy improvement over the standalone CNN.

\section{Error Analysis and Limitations}

While highly robust, the framework exhibited specific failure modes under extreme conditions, leading to the 1.06\% error margin on the private test set.

\subsection{Intra-Family Generator Confusion}
The primary vector for misclassification stems from the profound architectural homogeneity among generators within the same family (e.g., StableDiffusion3, StableDiffusion3.5, and StableDiffusionXL-Turbo), which utilize identical VAE decoders. While the semantic branch effectively delineates subtle stylistic variations in pristine conditions, severe degradations (e.g., heavy Gaussian blur) obliterate these high-frequency cues. Bereft of distinct semantic signals, the model structurally defaults to predicting the dominant generator family class, resulting in unavoidable intra-family confusion.

\subsection{Forensic Branch Failure under Compound Degradation}
While the forensic branch serves as a robust safety net, its efficacy decays when subjected to severe, compound degradations. The branch relies on the integrity of noise residuals extracted via high-pass filtering. However, operations such as AI super-resolution layered over severe JPEG compression fundamentally overwrite the intrinsic noise structure. Under such conditions, the extracted SVD spectral profile ceases to represent the original generator's fingerprint and instead maps to the degradation operator itself, effectively blinding the XGBoost classifier.

\section{Conclusion and Future Work}

In this study, we introduced a highly robust, dual-branch hybrid ensemble framework for the Synthetic Image Source Attribution Challenge. By synergistically fusing the semantic representational capacity of an EMA-regularized EfficientNet-B0 with the deterministic mathematical rigor of SVD spectral profiles and LBP micro-textures, our architecture effectively bridges the distribution shift between pristine training environments and heavily degraded deployment states. Achieving a 95.60\% private leaderboard accuracy using a strictly CPU-bound pipeline that trains and infers in under 6.5 hours establishes a highly scalable, resource-efficient paradigm for real-world deepfake detection.

Future research will focus on mitigating vulnerability to compound degradations. First, we aim to harden the semantic branch by injecting aggressive, multi-stage post-processing operations (e.g., recursive blurring combined with extreme quantization) directly into the data augmentation pipeline to force structural invariance. Second, transitioning from global SVD metrics to localized, patch-based spectral analyses could isolate pristine generator artifacts surviving in uncorrupted image sub-regions. Finally, we propose implementing a dynamic gating mechanism that explicitly calculates an image's structural degradation level at inference time, allowing the framework to algorithmically shift weight toward the spectral domain when spatial integrity is fully compromised.

\bibliographystyle{splncs04}

\end{document}